\documentclass{article}
\usepackage{algorithm}
\usepackage{algorithmic}
\usepackage{color} 

\usepackage[latin1]{inputenc}
\usepackage{amsmath}
\usepackage{amsfonts}
\usepackage{amssymb}
\usepackage{url}
\usepackage{enumitem}
\usepackage{graphics}

\def\Exp{{\mathbb{E}}}

\newcommand{\R}{{\mathbb R}}
\newcommand{\N}{{\mathbb N}}

\newtheorem{Theorem}{Theorem}
\newtheorem{Lemma}{Lemma}
\newtheorem{Postulate}{Postulate}

\title{Telling cause from effect\\ based on high-dimensional observations}

\author{
Dominik~Janzing$^1$, Patrik~O.~Hoyer$^2$,  Bernhard~Sch\"olkopf$^1$\\
${}$\\
{\small  1) Max  Plack Institute for Biological Cybernetics} \\
{\small T\"ubingen, Germany} \\
${}$\\
{\small 2) Helsinki Institute for Information Technology}\\
{\small University of Helsinki}\\ 
{\small Finland}
}

\date{September 24, 2009}

\begin{document}

\maketitle

\begin{abstract}
We describe a method for inferring linear causal relations
among multi-dimensional variables. The idea is to use an asymmetry between the distributions of cause and effect that occurs 
 if both the covariance matrix of the cause and the structure matrix mapping cause 
to the effect are independently chosen.
The method works for both stochastic and deterministic causal relations, provided that the dimensionality is sufficiently high (in some experiments, $5$ was enough).  It is applicable to Gaussian as well as non-Gaussian data.
\end{abstract}

\section{Motivation}

\label{Mot}

Inferring the causal relations that have generated statistical dependencies among a set of 
observed random  variables is challenging if no controlled randomized studies can be made.
Here, causal relations are represented as arrows connecting the variables,
and the structure to be inferred is a directed acyclic graph (DAG) \cite{Pearl:00,Spirtes}.
The constraint-based approach to causal discovery, one of the best known methods,
selects directed acyclic graphs that satisfy both the causal Markov condition and faithfulness: 
One accepts only those causal hypotheses that explain the observed dependencies and demand that all the observed {\it in}dependencies are 
imposed  by the
structure, i.e., common to all distributions that can be generated by the respective causal DAG.
However, the methods are fundamentally unable to distinguish between DAGs that induce the same set of  dependencies
(Markov-equivalent graphs).
Moreover, causal faithfulness is known to be violated if some  of the causal relations
are deterministic \cite{LemeireDiss}.  
Solving these problem requires reasonable prior assumptions, either implicitly or 
explicitly as priors on
conditional probabilities, as in Bayesian settings \cite{Heckerman1999a}.
However, the fact that deterministic dependencies exist in real-world settings shows that priors 
that are densities on the parameters of the Bayesian networks, as it is usually assumed, are problematic, and
the construction of good priors becomes difficult. 

Recently, several  methods have been  proposed that are able 
to distinguish between Markov-equivalent DAGs\footnote{In particular,
the elementary problem ``infer whether $X$ causes $Y$  or $Y$ causes $X$''  has been part of the challenge 
at the NIPS 2008  workshop ``Causality: Objectives and  Assessment'' \cite{MooijJanzingSchoelkopf08}}. 
Linear causal relations among non-Gaussian random variables can  be  inferred via independent-component-analysis (ICA) methods
 \cite{Kano2003,Shimizu2006}. 
The method of \cite{Hoyer} 
is able  to  infer causal directions
among real-valued variables 
if every effect is a  (possibly non-linear) function of its causes up to an additive noise term that
is independent of the causes. The work of \cite{ZhangUAI} augmented these models by applying a non-linear  function
after adding the noise term.
If the noise term vanishes or if all of the variables are Gaussian  and the relation is linear, all these methods fail.
Moreover, if the data are high-dimensional, the non-linear regression involved in the methods becomes hard to estimate.
 
Here we present a method that also works for these cases provided that the variables are multi-dimensional
with  sufficiently anisotropic covariance matrices.
The underlying idea is that the causal hypothesis $X\rightarrow Y$ is only acceptable if
the shortest description of the joint distribution $P(X,Y)$ is given by
separate descriptions of the input distribution $P(X)$ and the conditional distribution 
$P(Y|X)$ \cite{LemeireD}, expressing the fact that
they represent independent mechanisms of nature. \cite{Algorithmic} shows toy examples where 
such an independent choice often leads to joint distributions where $P(Y)$ and  $P(X|Y)$ 
satisfy non-generic relations indicating that $Y\rightarrow X$ is wrong.
Here we develop  this idea for the case of multi-dimensional variables $X$ and $Y$ with a  linear causal relation.
 
We  start with a motivating  example.
Assume that $X$ is a multivariate Gaussian variable with values in $\R^n$ 
and the isotropic covariance matrix $C_{XX}={\bf  I}$. 
Let $Y$ be another $\R^n$-valued variable that is deterministically influenced by $X$ via the linear 
relation $Y=AX$ for some $n\times n$-matrix $A$.
This induces the covariance matrix $$C_{YY}=A C_{XX} A^T=AA^T.$$ 
The converse causal hypothesis $Y\rightarrow X$  
becomes unlikely 
because $P(Y)$ (which is determined by the covariance matrix $AA^T$)
and $P(X|Y)$ (which is given by $X=A^{-1}Y$ with probability $1$)
are related  in a suspicious way, since the same matrix $A$ appears in both descriptions.
This untypical relationship between $P(Y)$ and $P(X|Y)$ can also be considered from the point of view of symmetries:
consider the set of covariance matrices $U C_{YY} U^T$ with $U\in O(n)$, where
$O(n)$ denotes the orthogonal group.
Among them, $C_{YY}$ is special because it is the only one that 
is transformed  into
the isotropic covariance matrix $C_{XX}$. More generally speaking, in light of the fact of how anisotropic   
the matrices 
\[
\tilde{C}_{XX}:=A^{-1}U C_{YY} U^T A^{-T}  
\]
are for  {\it generic} $U$, the hypothetical effect variable is surprisingly
isotropic for $U={\bf I}$ (here we have used the short notation $A^{-T}:=(A^{-1})^T$). 
We will show below that this remains true with high probability (in high dimensions) if we start with an
arbitrary  covariance matrix $C_{XX}$ and apply a random linear transformation $A$ chosen independently  of
$C_{XX}$. 

To understand why independent choices of $C_{XX}$ and $A$ typically induce 
untypical relations between $A^{-1}$ and $C_{YY}$ we also discuss the simple case that
 $C_{XX}$ and  $A$ are simultaneously diagonal with $c_j$ and  $a_j$ as 
corresponding diagonal entries. Thus $C_{YY}$ is also diagonal and its diagonal entries (eigenvalues)  are $a_j^2 c_j$.
We now assume that ``nature has chosen'' the 
values $c_j$ with $j=1,\dots,n$ independently from some distribution and $a_j$ from some other
distribution.
We can then interpret the values $c_j$  
as instances of $n$-fold sampling of the random variable $c$
with  expectation $\Exp(c)$
and the same for $a_j$. If we assume that 
$a$ and $c$ are  independent, we have
\begin{equation}\label{tracecomm}
\Exp(a^2c)= \Exp(a^2)\Exp (c)\,.
\end{equation}
Due to  the law of large numbers, this equation  will for large  $n$  approximatively be satisfied by the empirical averages, i.e.,
\begin{equation}
\frac{1}{n} \sum_{j=1}^n a^2_j c_j \approx  \left(\frac{1}{n} \sum_{j=1}^n a_j^2\right) \,\left(\frac{1}{n} \sum_{j=1}^n c_j\right)\,.
\end{equation}
For the backward direction $Y\rightarrow X$ we observe that the diagonal entries $\tilde{c}_j=c_j a_j^2$ of $C_{YY}$ 
and the diagonal entries $\tilde{a}_j=a_j^{-1}$ of $\tilde{A}:=A^{-1}$  have not been chosen independently
because
\begin{eqnarray*}
\Exp(\tilde{a}^2 \tilde{c})=\Exp(c)\,,
\end{eqnarray*}
whereas
\begin{eqnarray*}
\Exp(\tilde{a}^2)\Exp(\tilde{c}) &= & \Exp(a^{-2})\Exp(a^2 c)= \Exp(a^{-2})\Exp(a^2)\Exp(c)> \Exp(c) \,.
\end{eqnarray*}
The last inequality holds because the random variables $a^2$ and $a^{-2}$ are always negatively correlated
(this follows easily from the Cauchy-Schwarz inequality $\Exp(a^2)\Exp(a^{-2}) \leq  1$) except for the trivial case when they are constant. We thus observe a systematic violation of (\ref{tracecomm})
in the backward direction. The proof for non-diagonal matrices in Section~\ref{Iden} uses standard spectral theory, but is based upon the same idea.

The paper is structured as  follows. In Section~\ref{Iden}, 
we  
define an expression with traces on covariance  matrices and  show that
typical linear models induce backward models for which this expression attains values that would be untypical
for the forward direction. 
In Section~\ref{Exp} we describe an algorithm that is based upon this result  and discuss experiments
with simulated and real data. Section~\ref{Gen} proposes possible generalizations.

\section{Identifiability results}

\label{Iden}

Given a hypothetical causal model $Y=AX+E$ (where $X$ and $Y$  are $n$- and $m$-dimensional, respectively) 
we want to check whether the pair $(C_{XX},A)$ satisfies
some relation that typical  pairs $(UC_{XX}U^T,A)$ only satisfy with low probability if $U\in O(n)$ 
is randomly chosen.
To this end, we introduce  the renormalized trace 
$$\tau_n(.):={\tt  tr}(.)/n$$ 
for dimension $n$  and compare
the values 
\begin{equation}\label{tracecomp}
\tau_m (AC_{XX}A^T) \quad \hbox{ and } \quad \tau_n(C_{XX})\tau_m(AA^T)\,.
\end{equation}
One shows easily that the expectation of both values coincide if $C_{XX}$ is randomly drawn from
a distribution that is invariant under transformations $$C_{XX} \mapsto UC_{XX}U^T.$$  This is because
averaging the matrices $UC_{XX}U^T$ over all $U\in O(n)$  projects onto $\tau_n(C_{XX}) {\bf I}$  since
the average $U C_{XX} U^T$ commutes with all matrices and is therefore a multiple of the identity. 
For our purposes, it is decisive that  the typical case is close to this average, i.e.,
the two expressions in  (\ref{tracecomp}) almost coincide.
To show this,  
we need the following result \cite{Ledoux}:

\begin{Lemma}[L\'evy's Lemma]${}$\\ \label{Lev}
Let $g:S_n\rightarrow \R$ be 
a Lipschitz continuous function on  the $n$-dimensional sphere with 
\[
L:=\max_{\gamma\neq \gamma'} \frac{|g(\gamma)-g(\gamma')|}{\|\gamma-\gamma'\|}\,.
\]
If a point $\gamma$ 
on $S_n$ is randomly chosen according to an $O(n)$-invariant prior, it satisfies 
\[
|g(\gamma)-\bar{g}|\leq \epsilon 
\]
with probability at least $1-\exp(-\kappa (n-1)\epsilon^2/L^2)$
for some constant $\kappa$,  where $\bar{g}$ can be interpreted as the median or the average of $g(\gamma)$.
\end{Lemma}

Given the above Lemma, we can prove the following Theorem:

\begin{Theorem}[traces are  typically multiplicative]${}$\\
\label{Ind}  
Let $C$ be a symmetric, positive definite $n\times n$-matrix and  $A$ an arbitrary $m\times n$-matrix. 
Let $U$ be randomly chosen from $O(n)$ according to the unique $O(n)$-invariant distribution (i.e. the Haar measure).
Introducing the operator norm $$\|B\|:=\max_{\|x\|=1} \|Bx\|,$$
we have
\[
|\tau_m (AUCU^T  A^T)-\tau_n(C)\tau_m(AA^T)| \leq 2\epsilon  \|C\|\|AA^T\| 
\]
with probability at  least $q:=1-\exp(-\kappa (n-1)\epsilon^2)$
for some constant $\kappa$ (independent of  $C,A,n,m,\epsilon$).
\end{Theorem}

\noindent
Proof: for an arbitrary orthonormal system $(\psi_j)_{j=1,\dots,m}$ we have 
\[
\tau_m(AUCU^TA^T)=\frac{1}{m}\sum_{j=1}^m \langle \psi_j,AUCU^TA^T\psi_j\rangle \,.
\]
We define the  unit vectors $$\gamma_j:=U^TA^T\psi_j/\|A^T\psi_j\|.$$ Dropping the index $j$, we introduce
the function
\[
f(\gamma):=\langle \gamma, C \gamma\rangle\,.
\]
For a  randomly chosen $U\in O(n)$, $\gamma$ is a randomly chosen unit vector  according to a uniform prior
on  the $n$-dimensional sphere $S_n$.

The average of $f$ is given by 
$\bar{f}=\tau_n(C)$.
The Lipschitz constant is given by the operator norm of $C$, i.e.,
$
L=2 \|C\|\,.
$
An arbitrarily chosen $j$ satisfies
\[
|\langle \gamma_j, C \gamma_j\rangle -\tau_n(C)| \leq 2\epsilon  \|C\|
\]
with probability  $1-\exp(-\kappa (n-1)\epsilon^2)$. This follows from Lemma~\ref{Lev} after replacing $\epsilon$ with $\epsilon L$.  
Hence
\[
|\langle \psi_j,AU CU^TA^T \psi_j\rangle -\tau_n(C)\langle \psi_j, AA^T\psi_j\rangle |\leq 2 \epsilon \|C\|\|AA^T\|\,.
\]
Due to
\[
\tau_m(AUCU^TA^T) = \frac{1}{m}\sum_{j=1}^m \langle \psi_j ,A A^T\psi_j\rangle \langle \gamma_j, C\gamma_j\rangle \,,
\]
we thus have  
\[
|\tau_m(AUCU^TA^T)-\tau_m(AA^T)\tau_n(C)|\leq 2\epsilon  \|C\|\|AA^T\|\,.
\]
$\Box$

It is convenient to introduce
\[
\Delta(C,A):=\log \tau_m(ACA^T)-\log \tau_n(C)-\log  \tau_m(AA^T)
\]
as a scale-invariant measure for the strength of the violation of 
the equality of the expressions (\ref{tracecomp}).

We now restrict the attention to two special cases where we  can show that
$\Delta$ is non-zero for the backward direction.
First, we 
restrict the attention to deterministic models $$Y=AX$$
and the case that
 $m\geq n$ where $A$ has rank $n$.  This ensures that the backward model is also deterministic, i.e., 
\[ 
X=A^{-1}Y\,,
\]
with $(.)^{-1}$ denoting  the pseudo inverse.

The following theorem shows that 
$\Delta(C_{XX},A)=0$ implies $\Delta(C_{YY},A^{-1})<0$:   

\begin{Theorem}[systematic violation of trace multiplicativity]${}$\\
Let $n$ and $m$ denote the dimensions of $X$ and $Y$, respectively.
If $Y=AX$ and  $X=A^{-1}Y$, the covariance matrices satisfy
\begin{equation}\label{Delta}
\Delta (C_{XX},A) +\Delta(C_{YY},A^{-1})=-\log \left(1-{\tt Cov}(Z,1/Z)\right)+\log\frac{n}{m}\,,
\end{equation}
where $Z$ is a real-valued random variable whose distribution is the empirical distribution of eigenvalues
of $AA^T$, i.e., $\tau_m((AA^T)^k)=\Exp(Z^k)$ for all $k\in  \N$.
\end{Theorem}

\noindent
Proof: We have
\begin{equation}\label{fund}
\frac{\tau_n(C)}{\tau_m(ACA^T)\tau_n(A^{-1}A^{-T})}=\frac{1}{\tau_m(AA^T)\tau_n(A^{-1}A^{-T})} \frac{\tau_n(C)\tau_m(A^TA)}{\tau_m(ACA^T)}\,.
\end{equation}
Using $$\tau_n(A^{-1}A^{-T})=\tau_n(A^{-T}A^{-1})=\tau_n((AA^T)^{-1})=\frac{m}{n}\tau_m((AA^T)^{-1})$$ and
taking  the logarithm we obtain
\[
\Delta (ACA^T,A^{-1})= \log \frac{1}{\Exp(Z)\Exp(1/Z)}+\log \frac{n}{m} -\Delta (C,A)\,.
\]
Then the statement follows from
\[
{\tt Cov}(Z,1/Z)=1-\Exp(Z)\Exp(1/Z)\,.
\] 
$\Box$

Note that
 the term $-\log \left(1-{\tt Cov}(Z,1/Z)\right)$ in
eq.~(\ref{Delta}) 
will not converge to zero for dimension to infinity if the random matrices $A$ are drawn
in a way that ensures that the distribution of $Z$ converges to some distribution on $\R$ with non-zero variance.
Assuming this, $\Delta (C_{YY},A^{-1})$ tends to some negative  value if $\Delta (C_{XX},A)$ tends to zero for $n=m\to \infty$. 

We should, however, mention a problem that occurs for $m>n$ in the noise-less
case discussed here: Since  $C_{YY}$ has only rank $n$, we could equally well replace $A^{-1}$ with 
some other matrix $\hat{A}$ that coincides with $A^{-1}$ on  all of the observed $y$-values. 
For those matrices $\hat{A}$, the value $\Delta$ can get closer to zero because the term $\log n/m$ expresses
the fact that the image of $C_{YY}$ is orthogonal to the kernel of $A^{-1}$, which is already untypical 
for a generic model. 
 
It turns out that the  observed 
violation of the multiplicativity of traces can be interpreted in terms of relative entropy distances.
To show this,
we need the  following result:

\begin{Lemma}[relative entropy in terms of determinants and traces]${}$\\
Let $C$ be the covariance matrix of a centralized non-degenerate multi-variate Gaussian distribution $P_C$ in  $n$ dimensions.
Let the anisotropy  of $C$ be defined by the relative entropy distance to the closest isotropic Gaussian
\[
D(C):=\min_{Q \hbox{ isotropic }}D(P_C||Q)\,.
\] 
Then 
\begin{equation}\label{D}
D(C)=\frac{1}{2}\left(n\log \tau_n (C) -\log {\tt det}(C) \right)\,.
\end{equation}
\end{Lemma}

\noindent
Proof: the relative entropy distance of two centralized Gaussians with covariance matrices $C,C_0$ 
in $n$ dimensions is 
given by
\[
D(P_{C}||P_{C_0})=\frac{1}{2}\left(\log \Big(\frac{{\tt det} C_0}{{\tt det} C}\Big) +{\tt tr}(C_0^{-1} C) -n\right)\,.  
\] 
Setting $C_0=\lambda {\bf I}$, the distance is minimized for $\lambda=\tau_n(C)$, which yields
eq.~(\ref{D}). 
$\Box$

Straightforward computations show:

\begin{Theorem}[multiplicativity of traces and relative entropy]${}$\\
Let  $C$ and $A$ be $n\times n$-matrices with $C$ positive definite. 
Then
\[
D(ACA^T)=D(C)+D(AA^T)+\Delta(C,A)\,. 
\]
\end{Theorem}

Hence, for independently chosen $A$ and  $C$, the anisotropy of the  output covariance matrix
$ACA^T$ is approximately given by the anisotropy of $C$ plus the anisotropy of $AA^T$, which is the
anisotropy of  the output that $A$ induces on an isotropic input.   
For the backward direction, the anisotropy is smaller than the typical value.

We now discuss an example with a stochastic relation between $X$ and  $Y$.
We first consider the general linear model 
\[
Y=AX+E\,,
\]
where $A$ is  an  $n\times m$ matrix and $E$ is a noise term  (statistically independent of $X$) 
with covariance matrix $C_{EE}$.  We obtain
\[
C_{YY}=AC_{XX}A^T+C_{EE}\,.
\]
The corresponding backward model\footnote{For non-Gaussian $X,E$, this induces a joint distribution $P(X,Y)$  that  does not admit
a linear backward model 
with an {\it independent} noise $\tilde{E}$, we can then only obtain {\it uncorrelated} noise. 
We could in  principle already use this 
fact for causal inference \cite{Kano2003}. However, our method also works for the Gaussian case and 
if the dimension is too high for testing higher-order statistical dependences reliably.}
 reads
\[
X=\tilde{A}Y+\tilde{E}\,.
\]
with
\[
\tilde{A}:=C_{XY} C_{YY}^{-1}\,.
\]
Now we focus on the special case where $A$ is an orthogonal transformation and $E$ is isotropic, i.e.,
$C_{EE}=\lambda {\bf  I}$ with $\lambda>0$.
We then obtain a case where $C_{YY}$ and $\tilde{A}$ are related in a way that
makes $\Delta(C_{YY},\tilde{A})$ positive:

\begin{Lemma}[violation of multiplicativity of traces for a special noisy case]${}$\\
Let $Y=AX+E$
with  $A\in O(n)$ and 
the covariance matrix of $E$ be given by $C_{EE}=\lambda {\bf I}$.
Then we have
\[
\Delta(C_{YY},\tilde{A})>0\,.
\]
\end{Lemma}

\noindent
Proof: 
We have
\[
C_{YY}=ACA^T+\lambda  {\bf I} \quad \hbox{ and } \quad  C_{YX}=AC\,, 
\]
with $C:=C_{XX}$.
Therefore,
$$
\tilde{A}=CA^T(ACA^T+\lambda {\bf I})^{-1}=C (C+\lambda {\bf I})^{-1} A^T.
$$
One checks easily that the orthogonal transformation $A$ is irrelevant for the traces and we  thus  have  
\[
\Delta(C_{YY},\tilde{A})=\log \frac{\tau(C^2(C+\lambda {\bf I})^{-1})}{\tau(C+\lambda {\bf I}) \tau(C^2(C+\lambda {\bf I})^{-2})}=\log
\frac{\Exp\left(Z^2/(Z+\lambda)\right)}{\Exp(Z+\lambda)\Exp\left(Z^2/(Z+\lambda)^2\right)}\,,
\]
where $Z$ is a random variable of which distribution reflects  the distribution of eigenvalues of $C$.
The function $z\mapsto z/(z+\lambda)$ is monotonously increasing  for  positive $\lambda$ and $z$ and thus
also $z\mapsto z^2/(z+\lambda)^2$.   
Hence $Z+\lambda$ and  $Z^2/(Z+\lambda)^2$ are positively correlated, i.e., 
\[
\Exp(Z^2/(Z+\lambda))=\Exp((Z+\lambda) Z^2/(Z+\lambda)^2)> \Exp(Z+\lambda)\Exp\left(Z^2/(Z+\lambda)^2\right)\,,
\]
for all distributions of $Z$ with non-zero variance.
Hence the logarithm is positive and
thus $\Delta (C_{YY},\tilde{A})>0$.
$\Box$

Since the violation of the equality of the terms in (\ref{tracecomp}) can be  in both directions,
we propose to prefer the causal direction for which
$\Delta$ is closer to zero.

\section{Inference algorithm and experiments}

\label{Exp}

Motivated by the above theoretical results, we propose
to infer  the causal direction using
 Alg.~\ref{algtr}.\footnote{Please note that this algorithm, including the complete code to reproduce the experiments reported in this paper, is available as R code at: \texttt{http://www.cs.helsinki.fi/u/phoyer/code/hdlin.tar.gz}}

\begin{algorithm}[t]
   \caption{Identifying linear causal relations  via  traces}
   \label{algtr}
\begin{algorithmic}[1]
   \STATE {\bfseries Input:} $(x_1,y_1), \ldots, (x_k, y_k)$ \vspace{0.2cm}
   \STATE{ Compute the estimators $C_{XX}$, $C_{XY}$, $C_{YX}$, $C_{YY}$}
\STATE{Compute $A:=C_{YX} C_{XX}^{-1}$}
\STATE{Compute $\tilde{A}:=C_{XY}C_{YY}^{-1}$}
\IF{$|\log\tau_m(AC_{XX}A^T)-\log \tau_n(C_{XX})-\log \tau_m(AA^T)|>\epsilon + |\log\tau_n(\tilde{A}C_{YY}\tilde{A}^T)-\log \tau_m(C_{YY})-\log \tau_n(\tilde{A}\tilde{A}^T)|$}
\STATE{write ``$Y$ is the cause''}
\ELSE
\IF{$|\log\tau_n(\tilde{A}C_{YY}\tilde{A}^T)-\log \tau_m(C_{YY})-\log \tau_n(\tilde{A}\tilde{A}^T)|>\epsilon + |\log\tau_m(AC_{XX}A^T)-\log \tau_n(C_{XX})-\log \tau_m(AA^T)|$}
\STATE{write ``$X$ is the cause''}
\ELSE
\STATE{write ``cause cannot be  identified''}
\ENDIF
\ENDIF
\end{algorithmic}
\end{algorithm}

In light of the theoretical results, the following issues have  to be clarified by experiments with simulated data:

\begin{enumerate}
\item Is the limit for dimension to infinity already justified for moderate dimensions?
\item Is the multiplicativity of traces sufficiently violated for noisy models?
\end{enumerate}

Furthermore, the following issue has to be clarified by experiments with real data:

\begin{enumerate}[resume]
\item Is the behaviour of real causal structures qualitatively sufficiently close to our model with independent choices of $A$ and $C_{XX}$ according to a uniform prior?
\end{enumerate}

For the simulated data, we have generated random models $Y=AX+E$ as follows: We independently draw each element of the $m\times n$ structure matrix $A$ from a standardized Gaussian distribution. This implies that the distribution of column vectors as well as the distribution of row vectors
is isotropic. To generate a random covariance matrix $C_{XX}$, we similarly draw an $n\times n$ matrix $B$ and set $C_{XX}:=BB^T$. Due to the invariance of our decision rule with respect to the scaling of $A$ and $C_{XX}$, the structure matrix and the covariance can have the same scale without loss of generality. The  covariance $C_{EE}$ of the noise is generated in the same way, although with an adjustable parameter $\sigma$ governing the scaling of the noise with respect to the signal: $\sigma=0$ yields the deterministic setting, while $\sigma=1$ equates the power of the noise to that of the signal.

\begin{figure}[t]
{\large \hspace{9mm}{\sffamily {\bf a}\hspace{33mm}{\bf b}\hspace{33mm}{\bf c}\hspace{33mm}{\bf d}}}\\[-10mm]
\begin{center}
\resizebox{\textwidth}{!}{\includegraphics{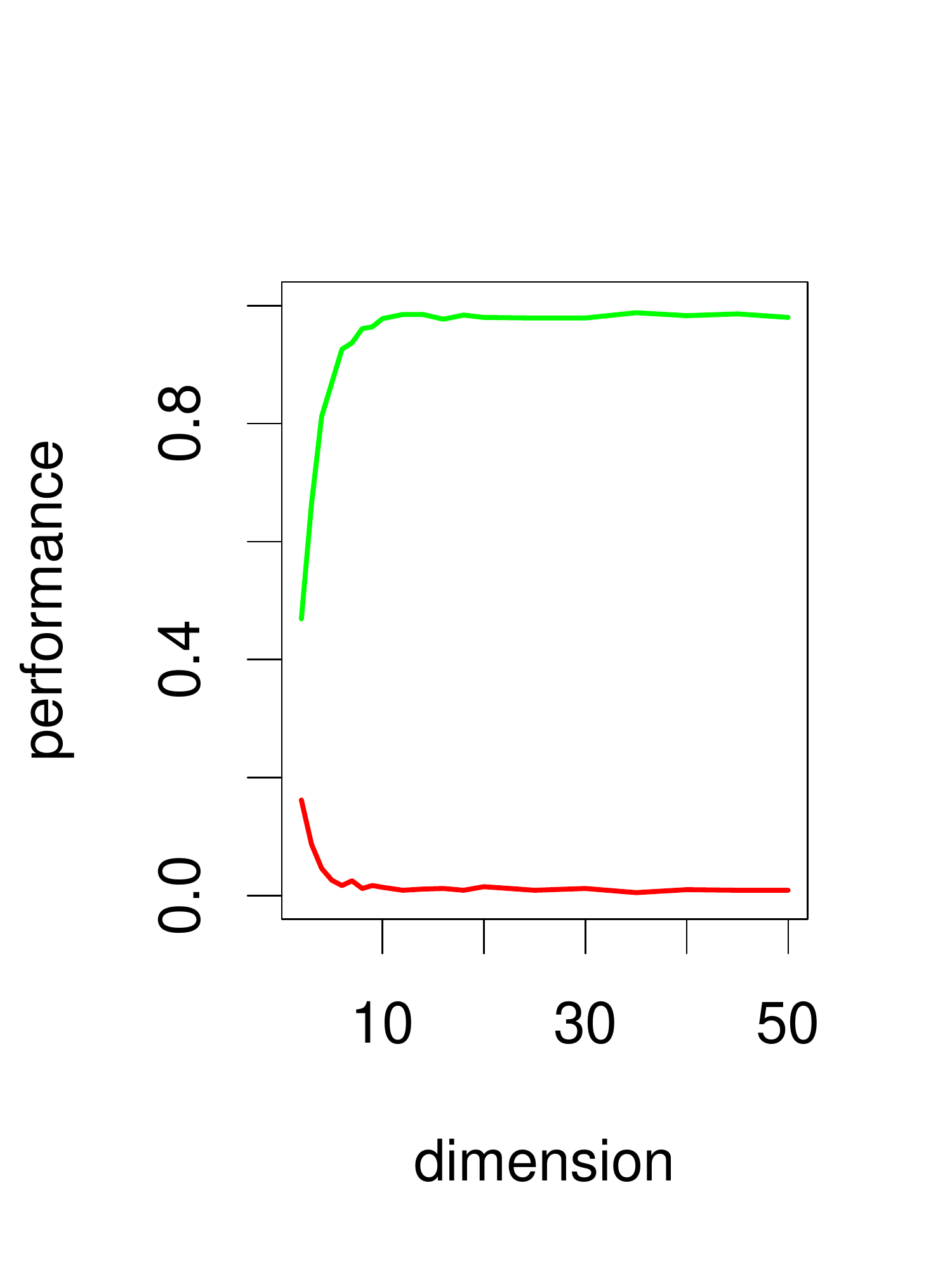}\includegraphics{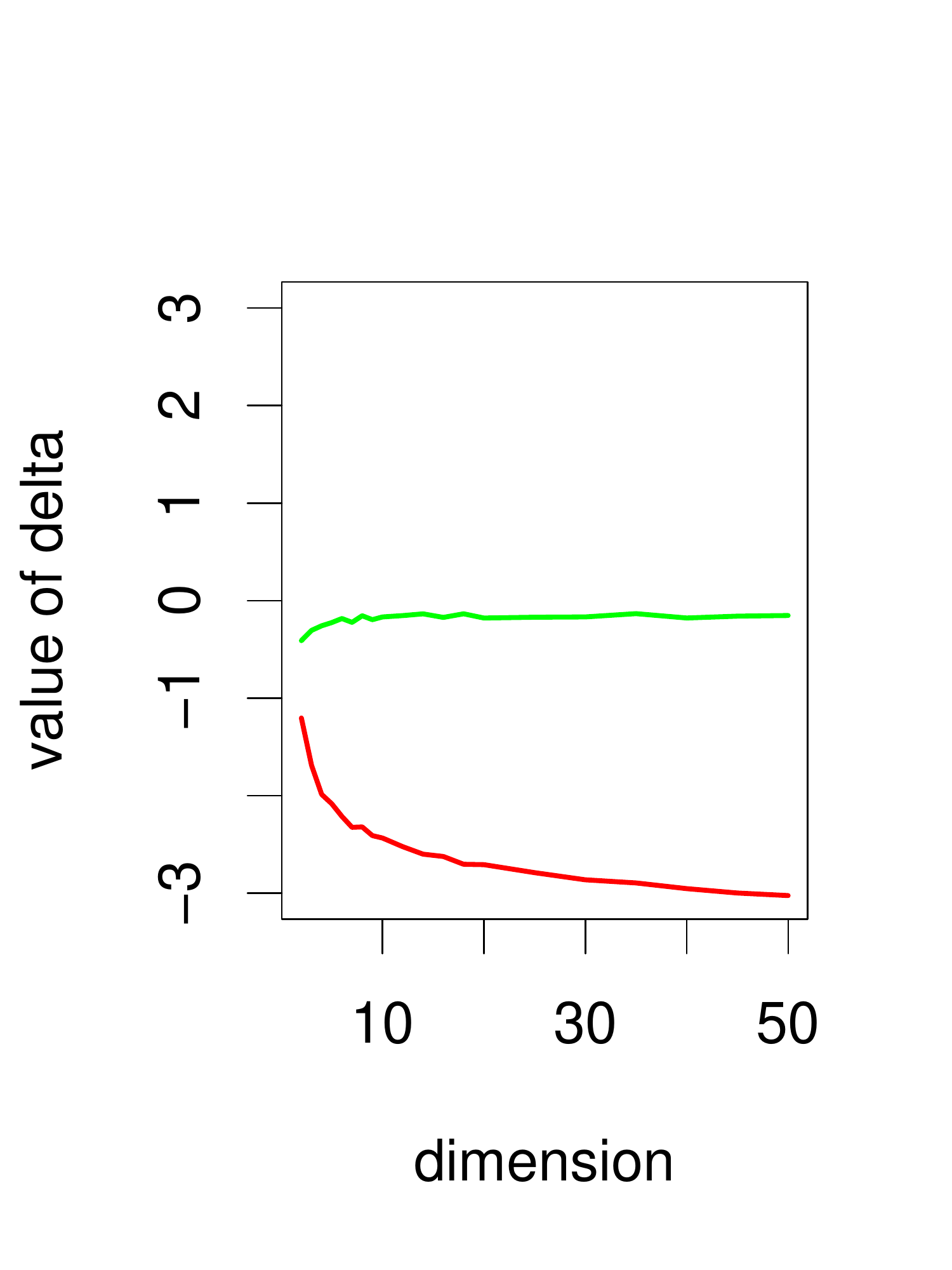}\includegraphics{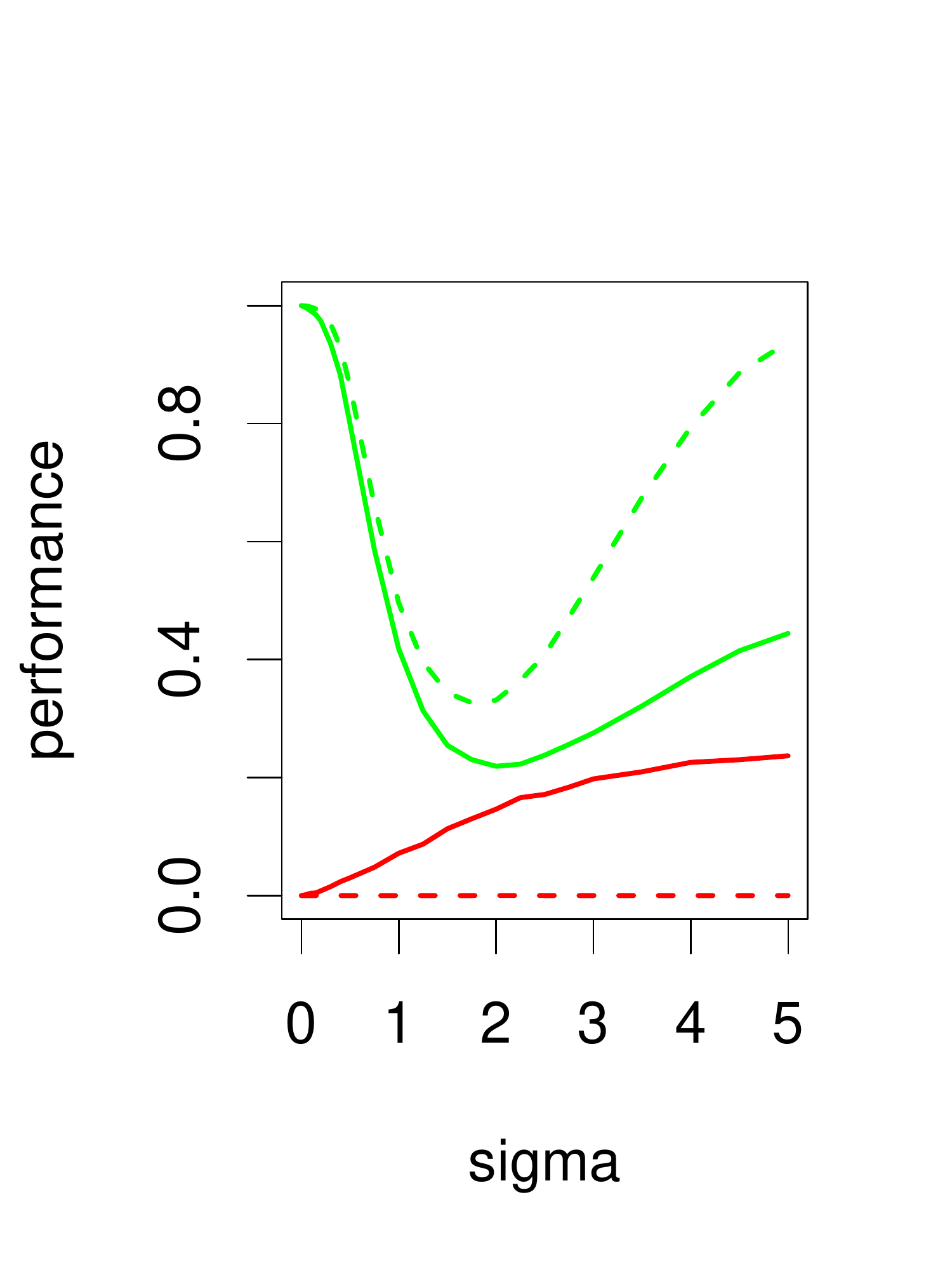}\includegraphics{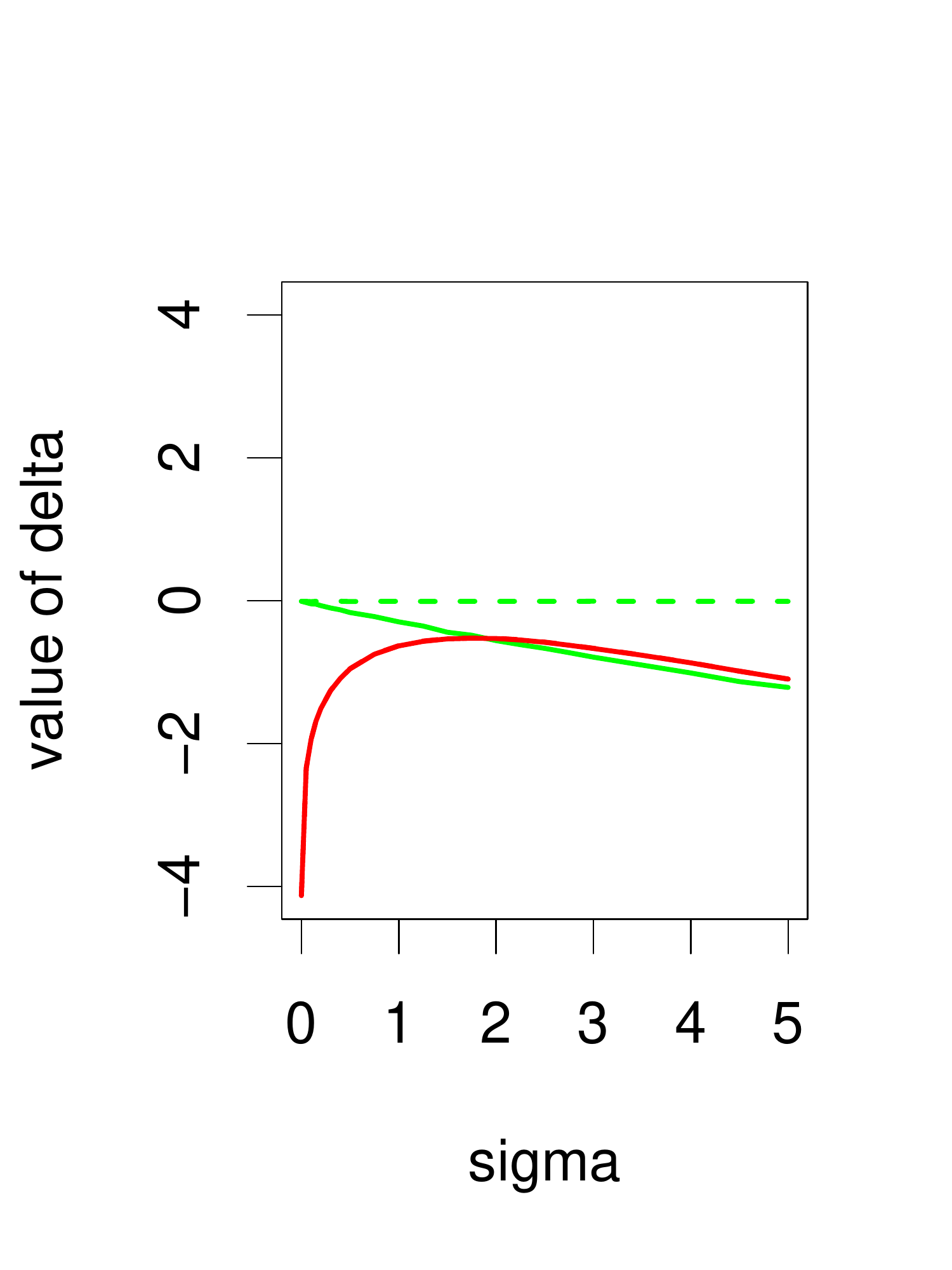}}
\end{center}
\caption{Simulation results. {\bf (a)} Performance of the method as a function of the input dimensionality $n$, when the output dimensionality $m=n$ and the sample size is $N=2n$. The green curve denotes the fraction of simulations on which the true causal direction was selected, while the red curve gives the fraction of wrong answers. {\bf (b)} Mean values of $\Delta$ corresponding to the true direction (green) vs the wrong direction (red). {\bf (c)} Performance as a function of noise level $\sigma$, for dimensionality $n=m=10$ and sample size $N=1000$. To compare, the dashed lines give the performance based on the exact covariance matrices rather than based on the samples. {\bf (d)} Mean values of $\Delta$ corresponding to the true direction (green) vs the wrong direction (red). See main text for discussion.
\label{fig:simulations}}
\end{figure}

First, we demonstrate the performance of the method in the close-to deterministic setting ($\sigma=0.05$) as a function of the dimensionality $n=m$ of the simulations, ranging from dimension 2 to 50. To show that the method is feasible even with a relatively small number of samples, we choose the number of samples $N$ to scale with the dimension as $N = 2n$. (Note that we must have $N \geq \min(n,m)$ to obtain invertible estimates of the covariance matrices.) The resulting proportion of correct vs wrong decisions is given in Fig.~\ref{fig:simulations}a, with the corresponding values of $\Delta$ in Fig.~\ref{fig:simulations}b. As can be seen, even at as few as 5 dimensions and 10 samples, the method is able to reliably identify the direction of causality in these simulations. 

To illustrate the degree to which identifiability is hampered by noise, the solid line in Fig.~\ref{fig:simulations}c gives the performance of the method for a fixed dimension ($n=m=10$) and fixed sample size ($N=1000$) as a function of the noise level $\sigma$. As can be seen, the performance drops markedly as $\sigma$ is increased. As soon as there is significantly more noise than signal (say, $\sigma > 2$), the number of samples is not sufficient to reliably estimate the required covariance matrices and hence the direction of causality. This is clear from looking at the much better performance of the method when based on the exact, true covariance matrices, given by the dashed lines. In Fig.~\ref{fig:simulations}d we show the corresponding values of $\Delta$, from which it is clear that the estimate based on the samples is quite biased for the forward direction. 

\begin{figure}[!t]
\begin{center}
\resizebox{\textwidth}{!}{\includegraphics{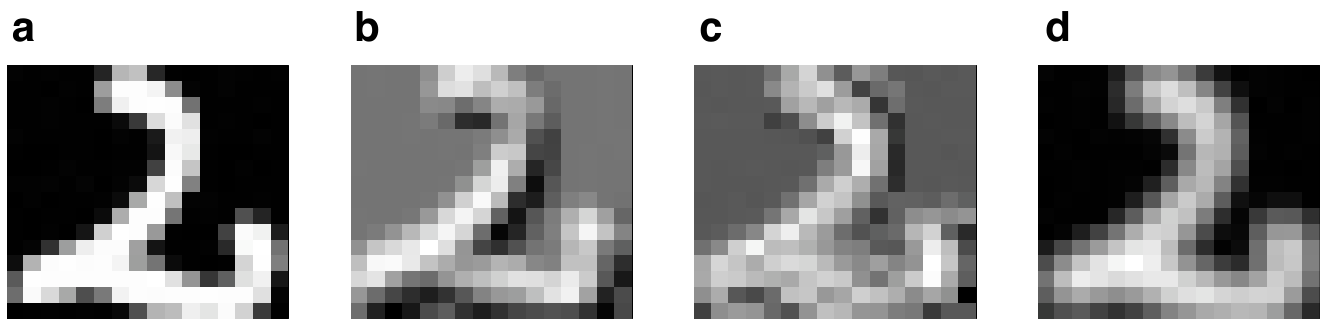}}
\end{center}
\caption{Example of original and filtered digit data. {\bf (a)} Original digit image. {\bf (b \& c)} Transformed image using two different random local translation-invariant linear filters. {\bf (d)} Transformed image using a simple local blur filter.
\label{fig:digits}}
\end{figure}

As experiments  with real data with known ground truth, we have chosen $16\times 16$ pixel images of handwritten digits \cite{LeCun90handwrittendigit}. As the linear map $A$ we have used both random local translation-invariant linear filters and also standard blurring of the images. (We added a small amount of noise to both original and processed images, to avoid problems with very close-to singular covariances.) See Fig.~\ref{fig:digits} for some example original and processed image pairs. The task is then: given a sample of pairs $(x_1,y_1)$ consisting of the picture $x_j$ and its processed counterpart $y_j$ infer which of the set of pictures $x$ or $y$ are the originals (`causes'). By partitioning the image set by the digit class (0-9), and by testing a variety of random filters (and the standard blur), we obtained a number of test cases to run our algorithm on. Out of the total of 100 tested cases, the method was able to correctly identify the set of original images 94 times, with 4 unknowns (i.e.\ only two falsely classified cases).

These simulations and experiments are quite preliminary and mainly serve to illustrate the theory developed in the paper. They point out at least one important issue for future work: the construction of unbiased estimators for the trace values or the $\Delta$. The systematic deviation of the  sample-based experiments from the covariance-matrix based experiments in Fig.~\ref{fig:simulations}c--d suggest that this could be a major improvement.

\section{Outlook: generalizations of the method} 

\label{Gen}

In  this section, we want to rephrase our theoretical results in a more abstract way to show the
general structure. We have rejected the causal hypothesis
$X\rightarrow Y$ if we observe that $\tau_n (AC_{XX}A^T)$ attains values that are not typical
among the set of  transformed input covariance matrices $UC_{XX}U^T$. 
In principle, we could have any function $K$  that maps the  output distribution
$P(Y)$ to some  value $K(P(Y))$.
Moreover, we could have any group $G$  of transformations
$g$ on the input variable $X$ that define transformed input distributions via
\[
P_g(X)=P(g^{-1}X)\,.
\]
Applying the conditional $P(Y|X)$ to $P_g(X)$ 
defines output distributions $P^{(g)}(Y)$ that we compare to $P(Y)$. 
In particular, we check whether 
the value $K(P(Y))$ is typical for the set $K(P^{(g)}(Y))_{g\in G}$. 

\begin{Postulate}[distribution of effect is typical  for the group orbit]${}$\\
Let $X$ and $Y$ be random variables with  joint distribution $P(X,Y)$
and $G$ be a group of transformations of the value set  of $X$.
Let $K(.)$ be some real-valued function on the  probability distributions of $Y$.
The causal hypothesis $X\rightarrow Y$ is unlikely if 
$K(P(Y))$ is smaller or greater than the big majority  of all distributions $(P^{(g)}(Y))_{g\in G}$ 
\end{Postulate}

Our prior knowledge about the structure of the data set determines the appropriate choice of $G$.
The idea is that $G$ expresses a set of transformations  that generate input distributions  $P_g(X)$
that we consider equally likely. 
The permutation 
of components of $X$ also defines an  interesting 
transformation group. For time series, the translation group would be the most natural choice. 

Interpreting this approach in a Bayesian way,
we thus use symmetry properties of priors without the need to explicitly define the priors themselves.

\section{Discussion}

Our experiments with simulated data suggest that the method performs quite well already for moderate dimensions provided that the noiselevel is not too high.
Certainly, 
the model of drawing $C_{XX}$  according to a distribution that  is invariant under $C_{XX} \mapsto UC_{XX}U^T$
may be inappropriate for many practical applications. 
However, as the example with diagonal matrices in Section~\ref{Mot}  
shows, the statement $\Delta(C_{XX},A)=0$ holds for a much broader class of models.
For this reason,
the method could also be  used as a sanity check for causal hypotheses among one-dimensional variables.
Assume, for instance, one has a causal DAG $G$ connecting $2n$ variables attaining values in  $\R$. 
If $X_1,\dots,X_{2n}$ is an ordering that is  consistent  with  $G$, we define
$Y:=(X_1,\dots,X_n)$ and $W:=(X_{n+1},\dots,X_{2n})$  and check the hypothesis $Y\rightarrow W$ using   our method.
Provided  that the true causal relations are linear, such a hypothesis should be accepted for every possible ordering
that  is consistent with the true causal DAG.
This way one  could, for instance, check the  causal relation between genes by clustering their
expression levels to vector-valued variables.


\end{document}